\documentclass[bdcc,article,accept,moreauthors]{Definitions/mdpi}

\firstpage{1}
\pubvolume{1}
\issuenum{1}
\articlenumber{0}
\pubyear{2026}
\copyrightyear{2026}
\externaleditor{ } 
\datereceived{5 June 2026}
\daterevised{20 August 2026} 
\dateaccepted{28 August 2026}
\datepublished{ }
\usepackage{xcolor}
\newcommand{\rev}[1]{#1} 

\Title{Measuring Reasoning Quality in LLMs: A Multi-Dimensional Behavioral Framework}

	\Author{Ali 
 Şenol $^{1,2,}$*\orcidA{}, Garima Agrawal $^{2,3}$\orcidB{} and Huan Liu $^{2}$\orcidC{}}

\AuthorNames{Ali Şenol, Garima Agrawal and Huan Liu}

\address{%
		$^{1}$ \quad Department of Computer Engineering, Tarsus University, Takbaş M. Kartaltepe S., \mbox{Tarsus 33400,~Mersin, Türkiye}\\
		$^{2}$ \quad School of Computing and Augmented Intelligence (SCAI), Arizona State University (ASU), \mbox{Tempe, AZ 85281, USA;} garima@humaconn.com (G.A.); huanliu@asu.edu (H.L.)\\
		$^{3}$  \quad HumaConn AI Consulting, \mbox{Queen Creek, AZ 85140, USA}}

	\corres{Correspondence: alisenol@tarsus.edu.tr}

\abstract{Despite remarkable progress on reasoning benchmarks, current LLM evaluation practice remains anchored to final-answer correctness. This provides limited insight into \emph{how 
} models reason, \emph{how reliably} they behave under contextual variation, or \emph{how efficiently} they reach conclusions. This paper proposes RQEval, a unified multi-dimensional framework for measuring LLM reasoning quality from a behavioral perspective. The framework operationalizes six theoretically grounded dimensions rooted in cognitive science: Correctness~(CQ), Consistency~(CS), Robustness~(RS), Local Logical Coherence~(LS), Efficiency~(ES), and Stability~(SS). It also introduces deployment-aware aggregation, enabling context-specific model selection beyond accuracy-based leaderboards. Applying RQEval to seven LLMs across four benchmarks reveals an \emph{outcome-level} cluster (CQ, CS, RS, and ES) characterized by strong intercorrelations and a \emph{trace-level} layer in which LS is empirically distinct from the outcome-level metrics, whereas SS retains moderate associations with several of them. Across the 28 model--dataset observations, LS showed no statistically significant correlation with any other dimension. Efficiency-weighted deployment scenarios, nevertheless, produced limited ranking inversions among models that were otherwise ranked consistently across weighting schemes. The resulting pipeline provides a foundation for diagnosing LLM reasoning behavior across deployment contexts, while highlighting domain-specific validation as an important direction for future work.}

\newcolumntype{L}[1]{>{\raggedright\arraybackslash}p{#1}}
\newcolumntype{C}[1]{>{\centering\arraybackslash}p{#1}}
\newcolumntype{R}[1]{>{\raggedleft\arraybackslash}p{#1}}

\begin{document}

\section{Introduction}
\label{sec:introduction}

LLMs have achieved remarkable success across
mathematical, logical, and commonsense reasoning tasks, fueling rapid
deployment in high-stakes domains such as clinical decision support, legal
analysis, automated scientific reasoning, and domain-specific
fraud detection~\cite{senol2026domain}. Yet the dominant evaluation
paradigm remains correctness-centric: a model is judged by whether its
final answer matches a ground-truth label, with little systematic attention
to \emph{how} it reasoned, \emph{how reliably} it behaves under contextual
variation, or \emph{how efficiently} it reaches correct conclusions.

This reductionism carries real costs. Cognitive science has established for
decades that reasoning quality is inherently multi-dimensional: a reliable
reasoner must produce accurate conclusions, maintain coherent inferential
chains, remain stable under equivalent re-formulations, and allocate
resources efficiently
\cite{simon1957models,stanovich2011rationality,thagard2000coherence,lieder2020resource}.
Collapsing these properties into a single accuracy score, therefore, discards
information critical for deployment decisions, particularly in
accountability-sensitive settings where the \emph{process} of reasoning
is subject to audit. This concern is not merely theoretical: recent
empirical work confirms that LLMs can generate plausible-sounding
reasoning chains causally disconnected from their final answers
\cite{lanham2023measuring,barez2025chain}, produce substantially different
outputs under semantically equivalent inputs~\cite{wang2022self,turpin2023language}, and exhibit systematic
cognitive-bias-like deviations that vary by model architecture
\cite{malberg2025comprehensive}.

We address this gap with three research questions:
\begin{itemize}
    \item \textbf{RQ1 
}: Can a
single unified framework simultaneously capture the correctness, coherence,
consistency, robustness, efficiency, and stability of LLM reasoning? These six dimensions are motivated by the cognitive science accounts of reasoning quality reviewed in Section~\ref{sec:related}.
    \item \textbf{RQ2}: Are these six dimensions empirically independent, or does accuracy subsume the others? 
    \item \textbf{RQ3}: Does deployment context materially change model rankings, and can the framework quantify these changes? 
\end{itemize}

\begin{figure}[H]
  \includegraphics[width=25em]{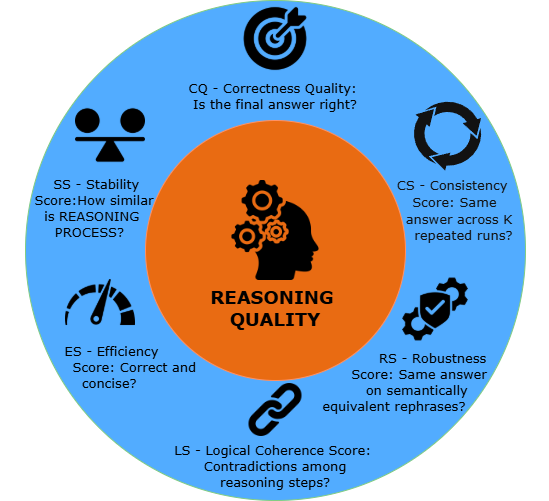}
  \caption{Conceptual 
 overview of the six reasoning-quality dimensions:
  Correctness (CQ), Consistency (CS), Robustness (RS), Local Logical
  Coherence (LS), Efficiency (ES), and Stability (SS).}
  \label{fig:reasoning}
\end{figure}

To answer these research questions, this paper makes four contributions:
\begin{enumerate}
    \item A theoretically grounded
six-dimensional (see Figure~\ref{fig:reasoning}) behavioral framework rationalizing cognitive science
principles as measurable LLM properties;
    \item Discriminant validity evidence showing that the six dimensions organize into two behavioral layers: a strongly intercorrelated outcome-level cluster (CQ, CS, RS, and ES) and a trace-level layer in which LS remains empirically orthogonal while SS is only partially distinct, with local logical coherence emerging as the framework's strongest orthogonal signal in the evaluated sample (Section~\ref{sec:discval});
    \item A systematic demonstration that multi-dimensional reasoning
profiles expose deployment-relevant ranking differences invisible to single-metric evaluation, something that is most pronounced among smaller open-weight models;
    \item RQEval, a reproducible, model-agnostic evaluation pipeline applicable to any LLM
without access to model weights or internal states.
\end{enumerate}

The rest of this paper is organized as follows. Section~\ref{sec:related} reviews related work on cognitive science foundations and chain-of-thought faithfulness, highlighting the limitations of current evaluation frameworks. Section~\ref{sec:methodology} details the proposed multi-dimensional framework, including metric definitions, aggregation strategies, and experimental setup. Section~\ref{sec:results} presents experimental results across seven LLMs and four benchmarks, with subsections on overall performance, per-dataset analysis, cross-cutting findings, deployment-aware model selection, and discriminant validity. Section~\ref{sec:discussion} discusses the implications of our findings for research and practice, and Section~\ref{sec:conclusion} concludes the paper.

\section{Related Work}
\label{sec:related}

\subsection{Cognitive Science Foundations}

Cognitive science provides the theoretical grounding for our framework.
Simon's bounded rationality framework \cite{simon1957models} established
that real reasoning systems operate under finite resources and must balance
accuracy with effort, directly motivating our Efficiency (ES) dimension.
Building on this foundation, Thagard's constraint-satisfaction model
\cite{thagard2000coherence} and Anderson et al.~\cite{simon2004integrated}
establish that high-quality reasoning additionally requires mutually
consistent inferential steps, not merely a correct conclusion,
grounding our Local Logical Coherence (LS) dimension. Extending the rationality
perspective to behavioral consistency, Stanovich \cite{stanovich2011rationality}
and Evans \cite{evans2007hypothetical} argue that rational agents should
produce consistent judgments across logically equivalent formulations,
motivating our Consistency (CS) dimension. Similarly, Kahneman and
Tversky \cite{kahneman1979prospect} and Gigerenzer and Selten
\cite{gigerenzer2001bounded} demonstrate that high-quality reasoning should
remain stable under minor representational changes, underpinning our
Robustness (RS) dimension. Lieder and Griffiths \cite{lieder2020resource}
further formalized resource-rational cognition, showing that good reasoners
optimize the tradeoff between correctness and computational cost. Together,
these accounts converge on a picture of reasoning as inherently
multi-dimensional, a view reinforced by Binz et al.~\cite{binz2025foundation}
and Bordalo \cite{bordalo2025cognitive}, who show that representational
stability and attentional consistency across tasks are essential
process-level criteria irreducible to accuracy alone.

\subsection{Chain-of-Thought Faithfulness and Robustness}

These theoretical criteria establish what high-quality reasoning should
exhibit; however, recent research on chain-of-thought prompting raises a
separate question of whether observable reasoning traces faithfully
reflect the processes that produce model outputs. The introduction of chain-of-thought prompting
\cite{wei2022chain,kojima2022large} raised expectations that LLM
reasoning traces could serve as interpretable records of model inference.
These expectations have been substantially qualified by a converging body
of evidence. Lanham et al.~\cite{lanham2023measuring} showed that CoT
traces frequently do not reflect the actual decision process; Barez et
al.~\cite{barez2025chain} formalize faithfulness in terms of procedural
soundness, causal relevance, and completeness, and find that current CoT
methods often fail all three. Paul et al.~\cite{paul2024making} further
demonstrate via causal mediation analysis that faithfulness improvements
through iterative refinement remain modest and task-dependent. From a data-distribution perspective, Zhao et al.~\cite{zhao2025mirage}
further argue that CoT reasoning functions as a structured inductive bias
learned from in-distribution training data; its effectiveness is,
therefore, constrained by the degree of distributional divergence between
training and test queries. This structural account reinforces the need
to evaluate reasoning behaviorally rather than assuming that CoT traces
faithfully represent a single underlying reasoning process. Taken
together, these findings caution against treating reasoning traces as
faithful records of inference, a limitation that directly motivates
our LS metric, which evaluates trace reliability through NLI-based
step-to-step contradiction detection rather than assuming fidelity.

The same body of work reveals parallel gaps in robustness and consistency
evaluation. Kumar and Mishra \cite{kumar2025robustness} identify a
persistent gap: most evaluations test at most one or two robustness
dimensions in isolation, leaving compounded fragilities undetected.
\mbox{Singh et al.~\cite{singh2024robustness}} confirm this concern, showing
significant performance degradation under compounded perturbations even
when mild perturbations are individually tolerated. At the deployment
level, Bogavelli et al.~\cite{bogavelli2026evaluating} show that prompt
format variations alone cause up to 40\% performance fluctuations in
enterprise settings. Crucially, Liu et al.~\cite{liu2025your} distinguish
output consistency from process stability, finding that larger models can
be unstable at the process level even when output-consistent, a
dissociation our framework operationalizes as independent CS and SS
dimensions, enabling their separation to be measured directly.

\subsection{Limitations of Current Frameworks}

Across this body of work, a consistent pattern emerges:
existing evaluation practice provides an incomplete and potentially
misleading account of LLM reasoning behavior. At the benchmark level,
Mondorf and
Plank \cite{mondorf2024beyond} and Wan et al.~\cite{wan-etal-2024-logicasker}
demonstrate that correctness-based benchmarks fail to distinguish genuine
reasoning from pattern recognition. At the level of the reasoning
process, Lyu et al.~\cite{lyu2023faithful} show that intermediate steps
may not causally contribute to final predictions, while Yang et
al.~\cite{yang2023rethinking} raise concerns about benchmark contamination,
suggesting that strong benchmark performance does not imply reliable
reasoning ability. The problem is not merely statistical: Turpin et
al.~\cite{turpin2023language} show that LLMs can produce explanations
logically coherent yet causally misaligned with their internal decision
processes. Most directly relevant to our work, Becerra-Monsalve et
al.~\cite{becerra2026multi} report only moderate NLI-based correlation
with human judgments ($\rho \leq 0.453$), identifying systematic
limitations in detecting higher-level reasoning errors. Collectively, these
findings point to the same diagnostic: domain-specific benchmarks
\cite{cobbe2021training,hendryckstest2021} target correctness in narrow formats,
robustness studies examine only a subset of perturbation types, and
faithfulness research requires manual intervention, leaving no unified
pipeline that systematically assesses all six dimensions identified by
cognitive science as constitutive of reasoning quality.

\textbf{Positioning. 
} Our framework is distinguished along three axes:
\emph{theory-grounded} (each dimension is defined with explicit reference
to a cognitive science principle); \emph{behavioral} (dimensions are
measured through perturbation, repetition, and trace-analysis rather than
introspective inspection); and \emph{deployment-aware} (a configurable
aggregation mechanism allows practitioners to derive evidence-based model
recommendations, a capability absent from all prior frameworks).

\rev{The comparison closest in spirit and scale to ours is HELM
\cite{liang2022holistic}, which, likewise, evaluates multiple LLMs across
several ``scenarios'' along several axes (accuracy, calibration,
robustness, fairness, and efficiency, among others) rather than a single
leaderboard score. HELM's robustness axis, however, is operationalized
as accuracy under a fixed set of input perturbations (invariance and
equivariance tests) rather than as reproducibility of the model's own
answer under semantic-preserving reformulation of the \emph{same} item, and
its efficiency metric is measured in wall-clock/FLOPs terms external to
the model's own output rather than as a correctness-efficiency tradeoff
internal to each response. Critically, HELM does not include any
measure of intermediate reasoning-trace coherence: its accuracy metrics
are computed exclusively from final answers, with no mechanism analogous
to LS for detecting step-to-step contradiction within a chain-of-thought
trace, and no analogue of CS's within-item, cross-resampling consistency
check. Our framework can be read as narrower in scope than HELM (it
targets reasoning behavior specifically, not the full range of LLM
capabilities HELM surveys) but deeper along the reasoning axis
specifically, and, unlike HELM's fixed per-scenario scoring, adds a
configurable deployment-aware aggregation layer that lets the same six
raw scores be recombined for different operational priorities without
re-running the underlying evaluation.} Table~\ref{tab:gap_analysis} summarizes how RQEval compares with existing evaluation methodologies.

\begin{table}[h]
\centering
\caption{Comparison of prior work against the proposed framework.
\checkmark\ = explicitly measured and \texttimes\ = not addressed.}
\label{tab:gap_analysis}
\resizebox{\textwidth}{!}{%
\begin{tabular}{L{4.9cm}C{1.2cm}C{1.2cm}C{1.2cm}C{1.2cm}C{1.2cm}C{1.2cm}}
\toprule
\textbf{Study} & \textbf{CQ} & \textbf{CS} & \textbf{RS} & \textbf{LS} & \textbf{ES} & \textbf{SS} \\
\midrule
Wang et al.~\cite{wang2022self}                 & \checkmark & \checkmark & \texttimes & \texttimes & \texttimes & \texttimes \\
Kumar \& Mishra~\cite{kumar2025robustness}       & \checkmark & \texttimes & \checkmark & \texttimes & \texttimes & \texttimes \\
Singh et al.~\cite{singh2024robustness}          & \checkmark & \texttimes & \checkmark & \texttimes & \texttimes & \texttimes \\
Bogavelli et al.~\cite{bogavelli2026evaluating}  & \checkmark & \texttimes & \checkmark & \texttimes & \texttimes & \texttimes \\
Liu et al.~\cite{liu2025your}                   & \checkmark & \checkmark & \texttimes & \texttimes & \texttimes & \checkmark \\
Mondorf \& Plank~\cite{mondorf2024beyond}        & \checkmark & \texttimes & \texttimes & \texttimes & \texttimes & \texttimes \\
Becerra-Monsalve et al.~\cite{becerra2026multi}  & \texttimes & \texttimes & \texttimes & \checkmark & \texttimes & \texttimes \\
Lanham et al.~\cite{lanham2023measuring}         & \texttimes & \texttimes & \checkmark & \checkmark & \texttimes & \texttimes \\
Lyu et al.~\cite{lyu2023faithful}               & \texttimes & \texttimes & \texttimes & \checkmark & \texttimes & \texttimes \\
Liang et al.~\cite{liang2022holistic} (HELM)    & \checkmark & \texttimes & \checkmark & \texttimes & \checkmark & \texttimes \\
\midrule
RQEval 
 (This work)                              & \checkmark & \checkmark & \checkmark & \checkmark & \checkmark & \checkmark \\
\bottomrule
\end{tabular}%
}
\end{table}

\section{Methodology}
\label{sec:methodology}

\subsection{Framework Overview}
\label{sec:overview}

The goal of this framework is to provide a comprehensive characterization of LLM reasoning quality that supports deployment decisions in high-stakes domains. Achieving this goal is inherently a multi-dimensional problem: decades of cognitive science research establish that reasoning quality cannot be reduced to a single property such as correctness, because a reliable reasoner must simultaneously produce accurate conclusions, maintain coherent inferential chains, remain stable under equivalent reformulations, and allocate computational resources efficiently~\cite{simon1957models,stanovich2011rationality,thagard2000coherence,lieder2020resource}. Collapsing these properties into a single score, therefore, discards information that is critical for deployment decisions.

The choice of six dimensions follows directly from this theoretical account. Correctness (CQ) captures epistemic accuracy~\cite{anderson1990adaptive,oaksford2007bayesian}; Consistency (CS) operationalizes rational invariance~\cite{stanovich2011rationality,evans2007hypothetical}; Robustness (RS) addresses stability under perturbation~\cite{kahneman1979prospect,gigerenzer2001bounded}; Local Logical Coherence (LS) reflects constraint-satisfaction in inferential chains~\cite{thagard2000coherence,simon2004integrated}; Efficiency (ES) embodies bounded rationality~\cite{simon1957models,lieder2020resource}; and Stability (SS) captures behavioral reliability across stochastic runs~\cite{liu2025your}. Together, these six properties constitute a theoretically motivated core
set rather than an exhaustive taxonomy of reasoning quality. \rev{Other
dimensions, such as calibration, uncertainty awareness, causal
faithfulness, and self-correction, are plausible extensions, and we make
no claim of completeness.} \rev{The six dimensions are conceptually distinct by
design, each motivated independently by the cognitive-science account
above; whether they are also \emph{empirically} independent is a
separate question we test rather than assume. The discriminant validity
analysis in Section~\ref{sec:discval} finds that they are not uniformly
independent: 6 of 15 dimension pairs show acceptable discriminant
separation ($|r|<0.50$), while the remaining pairs cluster into an
outcome-level group (CQ, CS, RS, and ES) that is strongly intercorrelated
once computed from a shared canonicalized answer representation. 

This two-layer organization is examined empirically in Section~\ref{sec:results}.

observation derived from the evaluated models and datasets, not an
assumption built into the framework's design; Local Logical Coherence
(LS) is the dimension that empirically remains most clearly orthogonal
to the rest.}

Reasoning quality is, thus, modeled as follows:
\[
Q = f(\mathrm{CQ}, \mathrm{CS}, \mathrm{RS}, \mathrm{LS}, \mathrm{ES}, \mathrm{SS}).
\]

We refer to this six-dimensional behavioral model as RQEval
(Reasoning Quality Evaluation) throughout the remainder of the paper.

\subsection{Metric Definitions}
\label{sec:metricdefs}

We next translate each of these theoretically motivated dimensions into
an operational~metric.

\rev{Table~\ref{tab:notation} defines all symbols used in the metric
formulas below. $K=3$ and $P=3$
 were chosen as a practical compromise
between estimation reliability and computational cost: each additional
repeated generation multiplies API cost and wall-clock time across seven
models, four datasets, and 975 items, and $K=3$ is the minimum value
that permits pairwise agreement to be computed as a non-degenerate
statistic (a single pair at $K=2$ provides only a binary match/mismatch
signal, with no way to distinguish a model that is occasionally
inconsistent from one that is consistently inconsistent). We view
$K=3$ as sufficient to detect substantial behavioral instability of the
magnitude observed in this study (Section~\ref{sec:results}), but not as
a precision instrument for fine-grained consistency estimation; larger
$K$ would reduce sampling variance in the CS and SS estimates and is a
direct avenue for future work under a larger compute budget.}

\begin{table}[H]
\centering
\caption{\rev{Notation used in metric definitions.}}
\label{tab:notation}
\rev{
\begin{tabular}{L{2.0cm}L{11.0cm}}
\toprule
\textbf{Symbol} & \textbf{Definition} \\
\midrule
$N$ & Number of evaluation items in the dataset \\
$y_i$ & Gold (ground-truth) answer for item $i$ \\
$\hat{y}_i$ & Model's predicted answer for item $i$, extracted from the
raw generation by the multi-strategy matching pipeline \\
$\mathbb{I}(\cdot)$ & Indicator function: 1 if the condition holds, 0 otherwise \\
$K$ & Number of independent generations per item ($K=3$), sampled at
temperature $=0.7$ \\
$\hat{y}_i^{(k)}$ & Model's answer for item $i$ on the $k$-th independent run \\
$P$ & Number of semantic-preserving perturbations per item ($P=3$) \\
$\hat{y}_i^{(p)}$ & Model's answer to the $p$-th perturbed version of item $i$ \\
$C$ & Set of items answered correctly on the unperturbed input:
$C = \{i : \hat{y}_i = y_i\}$ \\
$s_j$ & The $j$-th reasoning step (sentence) in a generated trace \\
$n_i$ & Number of reasoning steps in the trace for item $i$ \\
$\psi(s_j, s_{j+1})$ & Binary NLI contradiction indicator between
consecutive steps $s_j$ and $s_{j+1}$ (1 if the NLI classifier's top
label is \textit{contradiction}, 0 otherwise) \\
$t_i$ & Number of output tokens generated for item $i$ \\
$T_i$ & Min--max normalized token count over the evaluated set,
$T_i = (t_i - t_{\min}) / (t_{\max} - t_{\min})$, where $t_{\min}$ and
$t_{\max}$ are the minimum and maximum observed token counts in the
current evaluation run \\
$r_i^{(k)}$ & Full reasoning trace for item $i$ on run $k$ \\
\bottomrule
\end{tabular}
}
\end{table}

\textbf{Correctness} 

(CQ) measures final-answer accuracy: the epistemic
goal of reasoning as the production of conclusions that accurately
correspond to the underlying problem structure
\cite{anderson1990adaptive,oaksford2007bayesian}. \rev{Because LLM outputs
are often verbose (\textit{``John has 8 apples.''} rather than
\textit{``8''}), $\hat{y}_i$ is extracted from the raw generation using a
type-aware canonicalization procedure. The matching rule is selected from
the gold-answer type before comparison. For numeric answers, the pipeline
uses the value following a \texttt{\#\#\#\#} delimiter when present
(GSM8K's standard final-answer marker \cite{cobbe2021training}) and
otherwise extracts the final number in the generation; values are compared
within a tolerance of $10^{-3}$. Binary answers require an unambiguous
\texttt{yes}/\texttt{no} lexical signal.
Multiple-choice answers require a
structurally anchored option letter A--D, such as a letter following
\textit{``answer is''} or \textit{``correct option is''}, rather than an
arbitrary occurrence of that letter. Other free-text answers use normalized
string equality or containment after lowercasing, trailing-punctuation
removal, and whitespace collapse. Restricting containment to free-text
answers prevents short numeric and categorical labels from matching
incidentally inside unrelated tokens. If no applicable canonical answer can
be extracted, the response is scored as incorrect. The same canonicalization
procedure is used for CQ, CS, and RS. The full implementation is available
in the released code (Section~\ref{sec:setup}).}
\[
\mathrm{CQ} = \frac{1}{N} \sum_{i=1}^{N} \mathbb{I}\!\left(y_i = \hat{y}_i\right)
\]

\textbf{Consistency (CS)} 
evaluates output stability across $K=3$
independent responses per instance at temperature $= 0.7$:
\[
\mathrm{CS} = \frac{1}{N} \sum_{i=1}^{N}
     \frac{2}{K(K-1)}
     \sum_{k=1}^{K} \sum_{l=k+1}^{K}
     \mathbb{I}\!\left(\hat{y}_i^{(k)} = \hat{y}_i^{(l)}\right)
\]

\textbf{Robustness (RS)}
measures performance under $P = 3$
semantic-preserving perturbations generated via a fixed rule-based
pipeline: (i)~one WordNet synonym substitution among content words, using
the most frequent synset and accepting the best candidate only when its
cosine similarity under spaCy \texttt{en\_core\_web\_md} exceeds $0.85$;
(ii)~syntactic reordering with the same spaCy dependency parser, either by
moving a fronted subordinate clause or by swapping coordinated clauses
while preserving logical scope; and (iii)~surface paraphrasing through
English $\to$ French $\to$ English back-translation with
\texttt{facebook/nllb-200-distilled-600M}. A strategy that cannot produce a
distinct perturbation retains its fixed slot and contributes zero to the
$P=3$ denominator; the original question is never substituted for a failed
perturbation. Semantic preservation was assessed in a double-annotated
validation sample, yielding Cohen's $\kappa=0.91$.
RS is computed exclusively over originally correct instances $C$ to prevent trivially high scores for consistently-wrong models~\cite{kahneman1979prospect,gigerenzer2001bounded}:
\[
\mathrm{RS} = \frac{1}{|C|} \sum_{i \in C}
     \frac{1}{P} \sum_{p=1}^{P}
     \mathbb{I}\!\left(\hat{y}_i^{(p)} = y_i\right)
\]

By 
 construction, RS is downstream of CQ: it is undefined for
models with zero correctness and structurally correlated with
CQ across models. This dependency is by design---robustness is
only meaningful given a correct baseline---but practitioners
should treat CQ--RS and CQ--ES correlations as structural
rather than informative signals. For computational completeness, the
implementation returns $\text{RS}=0$ when $C$ is empty.

\textbf{Local Logical Coherence (LS)} 
evaluates step-to-step consistency of
reasoning traces using the \texttt{cross-encoder/nli-deberta-v3-small}
model to classify each consecutive step pair $s_j, s_{j+1}$ as
\textit{contradiction}, \textit{neutral}, or \textit{entailment}
\cite{thagard2000coherence,simon2004integrated}; $\psi$ is the resulting
binary contradiction indicator (top predicted label), not a probability.
Reasoning steps are obtained by splitting generated responses at period
boundaries; line breaks in locally generated responses are normalized before
this split. Consecutive pairs are supplied as
\texttt{premise [SEP] hypothesis}, with truncation at 512 tokens and a batch
size of 32. If a batched NLI call fails, the implementation assigns a
neutral label to the affected pairs; if the NLI model is unavailable, it falls back
to a documented lexical-negation heuristic.
Models producing single-sentence responses receive $\text{LS} = 1.0$ by
convention, since a single atomic step admits no internal contradiction by
definition. We acknowledge this is conservative: a single-sentence
response may avoid reasoning entirely, and high LS should not be
interpreted as evidence of deep reasoning soundness but rather as absence
of detected local contradiction.
\[
\mathrm{LS} = 1 - \frac{1}{N} \sum_{i=1}^{N}
          \frac{1}{n_i - 1}
          \sum_{j=1}^{n_i - 1} \psi(s_j,\, s_{j+1})
\]

\textbf{Efficiency (ES)} 
measures the tradeoff between correctness and
token cost via a harmonic mean, grounding the framework in bounded
rationality \cite{simon1957models,lieder2020resource}. Let $T_i$ be the
min--max normalized token count of item $i$'s output, relative to the
minimum and maximum token counts observed across the evaluated set for
that model (Table~\ref{tab:notation}):
\[
\mathrm{ES} = \frac{1}{N} \sum_{i=1}^{N}
     \frac{2 \cdot \mathrm{CQ}_i \cdot (1 - T_i)}{\mathrm{CQ}_i + (1 - T_i)}
\]

\textbf{Stability (SS)} 
measures the semantic similarity of reasoning
traces across $K=3$ runs via BERTScore F1, operationalizing Liu et
al.~\cite{liu2025your}'s finding that process stability and output
consistency are decoupled. For each pair of traces $(r_i^{(k)},
r_i^{(l)})$, the implementation uses BERTScore $\geq 0.3.13$ with
\texttt{distilbert-base-uncased}, English language settings, and no baseline
rescaling. The three pairwise comparisons induced by $K=3$ are computed in
a single batch and averaged. Missing or failed traces remain in the fixed
$K$ denominator and contribute zero; Jaccard similarity is used only as a
runtime fallback when BERTScore is~unavailable:
\[
\mathrm{SS} = \frac{1}{N} \sum_{i=1}^{N}
     \frac{2}{K(K-1)}
     \sum_{k=1}^{K} \sum_{l=k+1}^{K}
     \text{BERTScore}(r_i^{(k)},\, r_i^{(l)})
\]

For each evaluation item, the primary response is used to compute CQ, LS,
and ES and also serves as the first of the $K=3$ responses used for CS and
SS. Two additional independent responses complete the fixed-$K$ sample,
and three fixed perturbation slots are evaluated for RS. Final answers are
canonicalized using the same type-aware procedure for CQ, CS, and RS,
whereas LS and SS operate directly on the generated reasoning traces.
Finally, the six dimension scores are aggregated using either equal weights
or a deployment-specific weight vector.

\subsection{Aggregation Strategies}

Dimension scores are aggregated as a weighted average $Q_w = \sum_{d} w_d \cdot d$, where $d \in \{\mathrm{CQ}, \mathrm{CS}, \mathrm{RS}, \mathrm{LS}, \mathrm{ES}, \mathrm{SS}\}$. Seven built-in weighting schemes address the common deployment contexts shown in Table~\ref{tab:weights}:

\begin{table}[h]
\centering
\caption{Weight vectors for evaluation scenarios. All weights sum to 1.0.}
\label{tab:weights}
\resizebox{\textwidth}{!}{%
\begin{tabular}{L{4.2cm}C{1.4cm}C{1.4cm}C{1.4cm}C{1.4cm}C{1.4cm}C{1.4cm}}
\toprule
\textbf{Scenario} & \textbf{CQ} & \textbf{CS} & \textbf{RS} & \textbf{LS} & \textbf{ES} & \textbf{SS} \\
\midrule
Balanced           & 1/6 & 1/6 & 1/6 & 1/6 & 1/6 & 1/6 \\
Safety Priority    & 0.30  & 0.05  & 0.30  & 0.25  & 0.05  & 0.05  \\
Accuracy Priority  & 0.50  & 0.10  & 0.15  & 0.15  & 0.05  & 0.05  \\
Efficiency Priority& 0.20  & 0.15  & 0.15  & 0.10  & 0.30  & 0.10  \\
Medical Triage     & 0.40  & 0.05  & 0.30  & 0.20  & 0.03  & 0.02  \\
Legal/Compliance   & 0.15  & 0.25  & 0.20  & 0.35  & 0.03  & 0.02  \\
Edge Device/IoT    & 0.30  & 0.03  & 0.10  & 0.05  & 0.50  & 0.02  \\
\bottomrule
\end{tabular}%
}
\end{table}

These weight vectors represent theoretically motivated
illustrative defaults. Domain practitioners should
calibrate them against their specific operational
requirements and risk tolerance; the framework
supports arbitrary custom weight vectors via
the configuration file. \rev{Two design choices merit
explanation. In Safety Priority, LS (not CS) receives the second-largest
weight after CQ and RS: safety-critical review is primarily concerned
with whether a reasoning chain is internally free of contradiction
(the property LS measures) rather than whether the same surface answer
recurs across repeated sampling, which reflects generation-time
stochasticity more than audit risk. In Accuracy Priority, CQ is weighted
more heavily than CS (0.50 vs.\ 0.10) rather than comparably: our own
discriminant validity analysis (Section~\ref{sec:discval}) finds CQ and
CS to be strongly correlated ($r=0.928$) once both are computed under the
same canonicalized-answer protocol, so weighting them near-equally would
largely double-count a single underlying signal.}

\subsection{Experimental Setup}
\label{sec:setup}

The experimental design was constructed to capture variation along two
axes: model access and reasoning-task structure. The model panel includes
closed-source API systems, a large open-weight model accessed through an
API, and smaller locally executed open-weight models. This selection enables
comparisons across model scale, access conditions, and deployment
constraints. The benchmark suite covers open-ended arithmetic reasoning,
multiple-choice knowledge and reasoning, binary commonsense reasoning, and
controlled synthetic stress tests. Accordingly, the model-by-dataset design
evaluates whether the proposed dimensions remain informative across
heterogeneous answer formats and reasoning~demands.

Seven LLMs spanning closed-source API models and open-weight local models
were evaluated (Table~\ref{tab:models}). API models accessed via the ASU
CreateAI gateway (GPT-4o-mini, Claude-Haiku-4.5, Gemini-2.5-Flash,
LLaMA-3.1-70B) were queried at temperature $= 0.7$ with
\texttt{max\_tokens} $= 256$, with the exception of Gemini-2.5-Flash, for
which \texttt{max\_tokens} was increased to $1024$: this model's internal
reasoning consumes tokens against the same budget as the visible response,
and at $256$ tokens responses were systematically truncated before an
answer was produced (verified on held-out probes) and $1024$ tokens left
ample headroom (observed completions of 200--430 tokens) while imposing no
change to the extraction or scoring pipeline. A minimal neutral system
prompt (\textit{``You are a helpful assistant.''}) was supplied to all
gateway-routed models to suppress the gateway's own default template
injection, which otherwise prepended several hundred to several thousand
tokens of boilerplate context ahead of the query. DeepSeek-V3 was queried
directly via the DeepSeek API (temperature $= 0.7$,
\texttt{max\_tokens} $= 256$, and no system prompt required). Local models
(Qwen2.5-1.5B, Phi-2) were executed on an NVIDIA GTX~1650
(4~GB~VRAM). Phi-2 (2.7B parameters) was loaded in float16 and Qwen2.5-1.5B
(1.5B parameters) required 4-bit (NF4) quantization via
\texttt{bitsandbytes} to remain within the available VRAM budget
alongside the evaluation pipeline's other memory demands (tokenizer,
NLI and BERTScore models loaded for LS/SS computation). \rev{Phi-2's
float16 footprint (5.56~GB peak resident memory, measured via
\texttt{torch.cuda.max\_memory\_allocated()} during a diagnostic
generation pass) exceeds the GPU's 4.30~GB dedicated VRAM capacity by
approximately 1.3~GB. Execution, nonetheless, completed successfully,
which we attribute to Windows' WDDM driver-level memory virtualization:
a corresponding increase in Task Manager's Shared GPU Memory counter was
observed during model loading, consistent with transparent spillover of
the excess allocation into system RAM. This execution mode is
functionally comparable to, although slower than, native VRAM residency.
Nevertheless, because ES is computed from output token counts rather
than wall-clock latency, this difference in execution speed does not
affect the ES calculation, though it may partly explain the longer
wall-clock runtime observed for Phi-2 relative to Qwen2.5-1.5B during
evaluation.}

\rev{\textbf{Reproducibility note.} 
During reproduction of the original
experiments on this gateway, we identified and corrected two
methodological issues in the evaluation pipeline, which are detailed here for
transparency. First, the multi-strategy answer-extraction pipeline
(described under Correctness below) exhibited a short-gold substring
leak: for single-character or short numeric gold answers, strategy~(ii)
(substring containment) could match incidentally inside an unrelated
token (e.g., gold \texttt{``D''} matching the \texttt{d} in
\textit{``drastic''}; gold \texttt{``5''} matching inside \textit{``25''}).
This inflated CQ on MMLU and StrategyQA in particular. We replaced the
substring strategy, for short categorical golds, with type-aware
canonical matching: numeric answers are compared as parsed floating-point
values (respecting a \texttt{\#\#\#\#}-delimited final answer when
present); yes/no answers require an unambiguous lexical signal (e.g.,
negation markers) rather than raw substring presence and multiple-choice
answers require a structurally anchored option letter (e.g., following
\textit{``answer is''}, \textit{``correct option is''}, or as the first token
of the response) rather than any bare occurrence of the letter. Second,
Consistency (CS) was originally computed as full-text equality across the
$K$ runs, which collapsed to near-zero for any model that rephrased a
correct answer differently across runs (e.g., \textit{``The answer is
30.''} vs.\ \textit{``5 $\times$ 6 = 30''} were scored as inconsistent). CS
is now computed over the same canonicalized answers used for CQ, so that
paraphrases of the same underlying answer are scored as consistent. Both
corrections are implemented in the released code and were applied
uniformly to all seven models; all CQ, CS, RS, and ES values reported
below reflect the corrected protocol. LS and SS, which operate on
reasoning traces rather than extracted answers, are unaffected.}
\begin{table}[H]
\small
\caption{Evaluated models. Local models were loaded from the Hugging Face Hub.}
\label{tab:models}
\resizebox{\textwidth}{!}{%
\begin{tabular}{L{3.0cm}L{2.2cm}C{1.2cm}C{1.6cm}C{3.0cm}C{1.8cm}C{2.4cm}}
\toprule
\textbf{Model} & \textbf{Provider} & \textbf{Type} & \textbf{Params} & \textbf{Access} & \textbf{Quant.} & \textbf{Device} \\
\midrule
GPT-4o-mini        & OpenAI      & Closed & Undisclosed & API (ASU CreateAI)   & N/A     & Remote    \\
Claude-Haiku-4.5   & Anthropic   & Closed & Undisclosed & API (ASU CreateAI)   & N/A     & Remote    \\
DeepSeek-V3        & DeepSeek AI & Closed & Undisclosed & API (direct)         & N/A     & Remote    \\
Gemini-2.5-Flash   & Google      & Closed & Undisclosed & API (ASU CreateAI)   & N/A     & Remote    \\
LLaMA-3.1-70B      & Meta        & Open   & 70B         & API (ASU CreateAI)   & N/A     & Remote    \\
Qwen2.5-1.5B       & Alibaba     & Open   & 1.5B        & Hugging Face         & 4-bit (NF4) & Local GPU \\
Phi-2              & Microsoft   & Open   & 2.7B        & Hugging Face         & float16 & Local GPU \\
\bottomrule
\end{tabular}%
}
\end{table}
We evaluate across 975 items from four benchmarks: GSM8K 
\cite{cobbe2021training} (250 items; arithmetic word problems; and primarily
exercises CQ and ES), MMLU \cite{hendryckstest2021} (225 items from 9 reasoning subjects: logical fallacies, formal logic, abstract algebra, elementary/high-school/college mathematics, statistics, conceptual physics, and philosophy; and controlled environment for CS and SS),  StrategyQA \cite{geva-etal-2021-aristotle} (250 items; implicit multi-step commonsense reasoning; binary yes/no; and informative for LS), and a Synthetic dataset (250 items constructed by the authors:
100 arithmetic word problems with numerical variation, 75~adversarial instances embedding deliberate logical contradictions into otherwise valid premises, and 75 robustness probes pairing each item with two surface-level paraphrases, all 
designed to stress-test RS and CS. Items were manually reviewed for semantic validity and absence of overlap with GSM8K training splits). Items were sampled uniformly at random from each benchmark's designated test split using a fixed random seed (seed $= 42$), with no filtering beyond the subject categories listed above for MMLU.

Table~\ref{tab:synthetic_examples} provides representative
examples from each Synthetic subset.

\begin{table}[H]
\caption{Representative items from the Synthetic dataset.}
\label{tab:synthetic_examples}

\small
\begin{tabularx}{\textwidth}{p{2.5cm}p{8.5cm}p{2cm}}
\toprule
\textbf{Type} & \textbf{Example Item} & \textbf{Target Dim.} \\
\midrule
Arithmetic & ``A train travels 60 km/h for 2.5 h. How far does it travel?'' & CQ, ES \\
\midrule
Adversarial & ``All mammals are warm-blooded. Whales are fish. Are whales warm-blooded?'' (deliberate contradiction in premise) & RS, CS \\
\midrule
Robustness probe & ``What is 144 divided by 12?''/``How much is 144 over 12?''/``Divide one hundred and forty-four by twelve.'' & RS \\
\bottomrule
\end{tabularx}
\end{table}

\section{Results}
\label{sec:results}

All six dimension scores range from 0 to 1, with higher values indicating
better performance. Unless otherwise stated, overall scores pool the 975
evaluation items across all four datasets. Because several outcome-level
metrics share the same canonicalized answer representation or correctness
signal, differences across dimensions should be interpreted as behavioral
profiles rather than as statistically independent effects.

\subsection{Overall Multi-Dimensional Reasoning Quality}

Table~\ref{tab:overall} shows that Claude-Haiku-4.5 achieves the highest performance across all aggregation
strategies ($Q_{\text{bal}}=0.808$), driven by superior CQ ($0.783$) and
RS ($0.895$). The value of multi-dimensional profiling remains apparent
even though, under the corrected protocol, aggregate rankings are highly stable across weighting schemes (see Section~\ref{sec:deployment}): DeepSeek-V3 and Gemini-2.5-Flash show similar balanced scores (0.766 and 0.756) yet reach them through different
profiles; DeepSeek is strongest on RS (0.927) and weakest on ES (0.540),
while Gemini is comparatively balanced across CQ, CS, and RS but leads on
LS (0.782). Equally striking, LLaMA-3.1-70B
($Q_{\text{bal}}=0.723$), despite being a 70B-parameter model, scores
below Claude-Haiku-4.5 and only marginally above GPT-4o-mini
($Q_{\text{bal}}=0.722$), suggesting that model scale alone does not
determine multi-dimensional reasoning quality.

\begin{table}[h]
\centering
\caption{Overall reasoning quality scores across all 975 items, computed
under the corrected canonical-answer-extraction protocol
(Section~\ref{sec:setup}). $Q_{\text{bal}}$ = balanced aggregate;
$Q_{\text{saf}}$ = safety-priority; $Q_{\text{acc}}$ = accuracy-priority; and
$Q_{\text{eff}}$ = efficiency-priority. All scores range from 0 to 1,
with higher values indicating better performance. The highest value in
each column is shown in bold.}
\label{tab:overall}
\resizebox{\textwidth}{!}{%
\begin{tabular}{L{3.2cm}C{0.8cm}C{0.8cm}C{0.8cm}C{0.8cm}C{0.8cm}C{0.8cm}
                C{1.0cm}C{1.0cm}C{1.0cm}C{1.0cm}}
\toprule
\textbf{Model} &
\textbf{CQ} & \textbf{CS} & \textbf{RS} & \textbf{LS} & \textbf{ES} & \textbf{SS} &
\textbf{$\mathbf{Q_{\text{bal}}}$} & \textbf{$\mathbf{Q_{\text{saf}}}$} & \textbf{$\mathbf{Q_{\text{acc}}}$} & \textbf{$\mathbf{Q_{\text{eff}}}$} \\
\midrule
GPT-4o-mini       & 0.600 & 0.646 & 0.877 & 0.811 & 0.465 & \textbf{0.930} & 0.722 & 0.748 & 0.688 & 0.662 \\
Claude-Haiku-4.5  & \textbf{0.783} & \textbf{0.781} & 0.895 & 0.836 & \textbf{0.643} & 0.910 & \textbf{0.808} & \textbf{0.829} & \textbf{0.807} & \textbf{0.775} \\
DeepSeek-V3       & 0.719 & 0.740 & \textbf{0.927} & 0.759 & 0.540 & 0.911 & 0.766 & 0.793 & 0.759 & 0.723 \\
Gemini-2.5-Flash  & 0.674 & 0.679 & 0.914 & 0.782 & 0.582 & 0.906 & 0.756 & 0.780 & 0.734 & 0.717 \\
LLaMA-3.1-70B     & 0.636 & 0.619 & 0.865 & 0.838 & 0.486 & 0.893 & 0.723 & 0.760 & 0.704 & 0.669 \\
Qwen2.5-1.5B      & 0.468 & 0.543 & 0.793 & 0.826 & 0.328 & 0.876 & 0.639 & 0.672 & 0.591 & 0.562 \\
Phi-2             & 0.426 & 0.490 & 0.609 & \textbf{0.860} & 0.395 & 0.823 & 0.601 & 0.611 & 0.543 & 0.537 \\
\bottomrule
\end{tabular}%
}
\end{table}

\begin{figure}[H]
  \includegraphics[width=\textwidth]{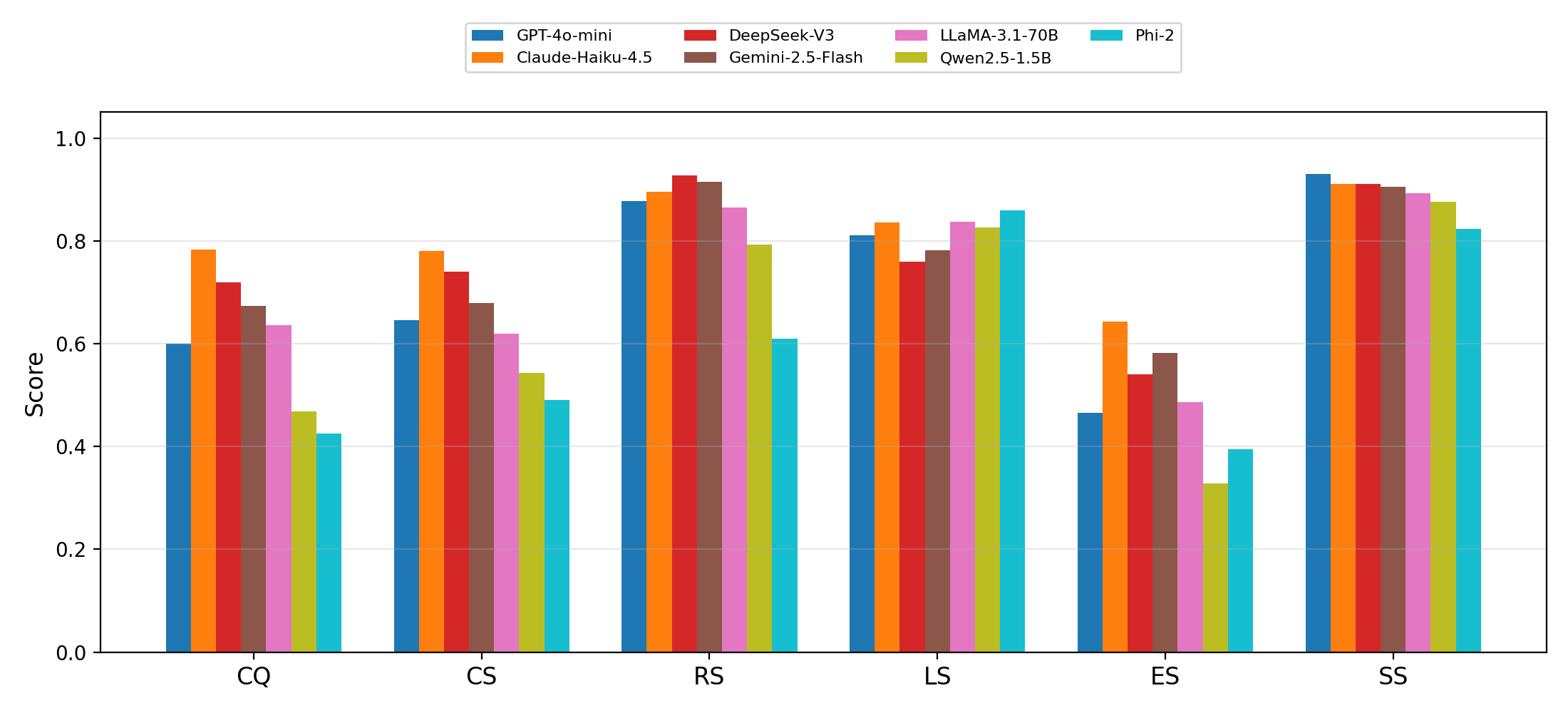}
  \caption{Comparative 
 scores for the six reasoning-quality dimensions
  across seven models, pooled over 975 items from four datasets. All scores
  range from 0 to 1, with higher values indicating better performance.
  CQ = Correctness; CS = Consistency; RS = Robustness; LS = Local Logical
  Coherence; ES = Efficiency; and SS = Stability.}
  \label{fig:bar_overall}
\end{figure}

A clear structural pattern emerges at the dimension level. Because CS is
computed over the same canonicalized answers as CQ (Section~\ref{sec:setup}), the two dimensions track each other closely across
models ($r=0.928$; Section~\ref{sec:discval}) rather than dissociating:
models that produce cleanly extractable answers tend to reproduce them
across the $K=3$ runs. Nevertheless, SS remains uniformly higher than CS
for every model (e.g., Claude-Haiku-4.5: $\text{SS}=0.910$
vs.\ $\text{CS}=0.781$; Phi-2: $\text{SS}=0.823$ vs.\ $\text{CS}=0.490$),
indicating that the \emph{semantic content} of reasoning traces is more
stable across runs than the \emph{exact extracted answer},
a finer-grained dissociation between semantic and surface-level consistency. This pattern extends to small models:
Phi-2 achieves the highest LS in the panel (0.860) despite the lowest CQ
(0.426). While this is consistent with the interpretation
that coherence is partially independent of correctness (LS--CQ:
$r=-0.083$, \textit{ns}; Section~\ref{sec:discval}), we note that the
NLI-based LS metric detects local step-to-step contradiction rather than
global reasoning validity; these scores should, therefore, be read as
evidence of local inferential consistency rather than deep semantic
soundness. Figure~\ref{fig:bar_overall} visualizes these dimension-level profiles across the seven models.

\begin{figure}[H]
  \includegraphics[width=0.72\textwidth]{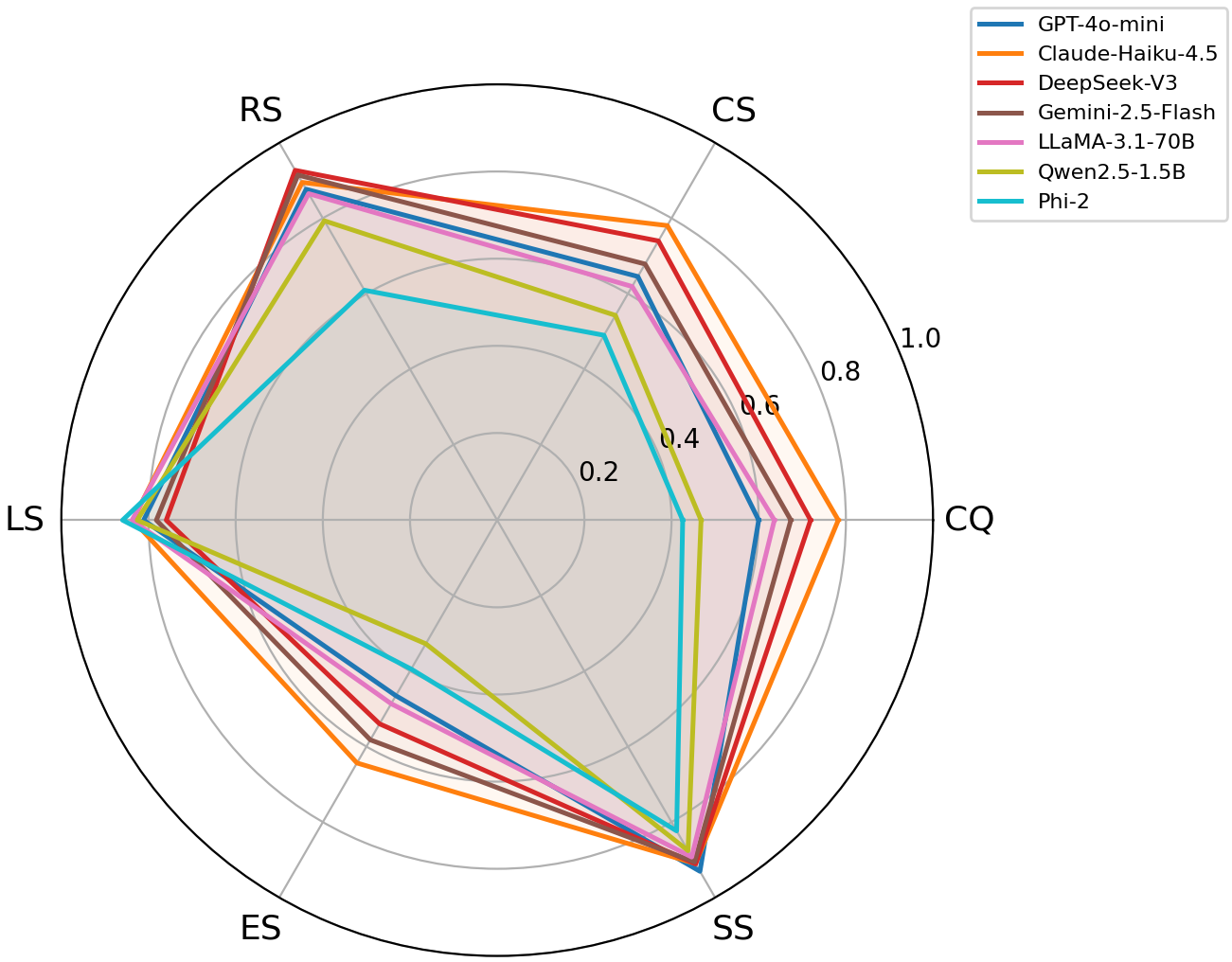}
  \caption{Radar 
plot of overall reasoning quality across the six
dimensions. Claude-Haiku-4.5 has the highest balanced aggregate score,
  whereas DeepSeek-V3's LS score is comparatively lower than its CQ, CS,
  and RS scores. All axes range from 0 to 1, with higher values indicating
  better performance.}
  \label{fig:radar_overall}
\end{figure}

\vspace{-6pt}
Figure~\ref{fig:radar_overall} provides the same overall comparison as a radar plot to emphasize cross-dimensional profile shapes.

\subsection{Per-Dataset Analysis}
\label{sec:perdataset}

Although the pooled results reveal broad model-level patterns, they
conceal substantial variation across benchmarks. Table~\ref{tab:perdataset}, therefore, reports the results separately for
each~dataset.

\textbf{GSM8K.} 
CQ ranges widely (0.240--0.904), with DeepSeek-V3 leading
(0.904), narrowly ahead of Claude-Haiku-4.5 (0.880); both substantially
outperform the remaining API models. \rev{Under the corrected extraction
protocol, GSM8K CQ is markedly lower for the two smallest models because
the numeric-extraction strategy credits only the final number in
each response.} Both small models maintain
relatively high LS given their CQ (Qwen2.5-1.5B: 0.713, Phi-2: 0.762),
suggesting that relatively high local coherence can coexist with low
final-answer correctness.

\textbf{MMLU.} 
\rev{Correctness is comparatively low for Gemini-2.5-Flash
($\text{CQ}=0.449$), which records the lowest CQ among the API-based
models on this dataset. As discussed in Section~\ref{sec:setup}, the
corrected extraction pipeline prevents spurious matches between
short gold labels and incidental occurrences of single option
letters.}
Claude-Haiku-4.5 remains the strongest ($\text{CQ}=0.924$). The constrained
\{A,B,C,D\} format still yields comparatively high RS values
(0.578--0.946) relative to the open-ended datasets.

\begin{table}[H]
\caption{Per-dataset breakdown for all seven evaluated models, computed
under the corrected canonical-answer-extraction protocol.}
\label{tab:perdataset}
\resizebox{\textwidth}{!}{%
\begin{tabular}{L{3.2cm}L{1.8cm}C{0.7cm}C{0.7cm}C{0.7cm}C{0.7cm}C{0.7cm}C{0.7cm}C{0.9cm}}
\toprule
\textbf{Model} & \textbf{Dataset} & \textbf{CQ} & \textbf{CS} & \textbf{RS} & \textbf{LS} & \textbf{ES} & \textbf{SS} & \textbf{$\mathbf{Q_{\text{bal}}}$} \\
\midrule
GPT-4o-mini      & GSM8K      & 0.552 & 0.567 & 0.804 & 0.670 & 0.343 & 0.931 & 0.645 \\
GPT-4o-mini      & MMLU       & 0.564 & 0.670 & 0.937 & 0.707 & 0.497 & 0.951 & 0.721 \\
GPT-4o-mini      & StrategyQA & 0.480 & 0.531 & 0.908 & 0.927 & 0.382 & 0.891 & 0.687 \\
GPT-4o-mini      & Synthetic  & 0.800 & 0.817 & 0.868 & 0.930 & 0.662 & 0.951 & 0.838 \\
\midrule
Claude-Haiku-4.5 & GSM8K      & 0.880 & 0.901 & 0.933 & 0.865 & 0.738 & 0.951 & 0.878 \\
Claude-Haiku-4.5 & MMLU       & 0.924 & 0.938 & 0.946 & 0.703 & 0.706 & 0.903 & 0.854 \\
Claude-Haiku-4.5 & StrategyQA & 0.624 & 0.623 & 0.897 & 0.871 & 0.416 & 0.865 & 0.716 \\
Claude-Haiku-4.5 & Synthetic  & 0.716 & 0.676 & 0.749 & 0.891 & 0.553 & 0.921 & 0.751 \\
\midrule
DeepSeek-V3      & GSM8K      & 0.904 & 0.924 & 0.975 & 0.680 & 0.694 & 0.927 & 0.851 \\
DeepSeek-V3      & MMLU       & 0.636 & 0.628 & 0.932 & 0.668 & 0.554 & 0.933 & 0.725 \\
DeepSeek-V3      & StrategyQA & 0.624 & 0.660 & 0.844 & 0.822 & 0.284 & 0.865 & 0.683 \\
DeepSeek-V3      & Synthetic  & 0.704 & 0.736 & 0.935 & 0.859 & 0.635 & 0.921 & 0.798 \\
\midrule
Gemini-2.5-Flash & GSM8K      & 0.812 & 0.861 & 0.951 & 0.672 & 0.622 & 0.948 & 0.811 \\
Gemini-2.5-Flash & MMLU       & 0.449 & 0.403 & 0.845 & 0.741 & 0.376 & 0.876 & 0.615 \\
Gemini-2.5-Flash & StrategyQA & 0.704 & 0.740 & 0.941 & 0.813 & 0.606 & 0.866 & 0.778 \\
Gemini-2.5-Flash & Synthetic  & 0.708 & 0.683 & 0.880 & 0.898 & 0.655 & 0.930 & 0.792 \\
\midrule
LLaMA-3.1-70B    & GSM8K      & 0.844 & 0.812 & 0.872 & 0.721 & 0.647 & 0.913 & 0.802 \\
LLaMA-3.1-70B    & MMLU       & 0.671 & 0.690 & 0.903 & 0.801 & 0.553 & 0.868 & 0.748 \\
LLaMA-3.1-70B    & StrategyQA & 0.368 & 0.287 & 0.783 & 0.891 & 0.220 & 0.858 & 0.568 \\
LLaMA-3.1-70B    & Synthetic  & 0.664 & 0.695 & 0.868 & 0.935 & 0.526 & 0.929 & 0.769 \\
\midrule
Qwen2.5-1.5B     & GSM8K      & 0.344 & 0.312 & 0.713 & 0.713 & 0.195 & 0.906 & 0.530 \\
Qwen2.5-1.5B     & MMLU       & 0.382 & 0.539 & 0.876 & 0.852 & 0.361 & 0.892 & 0.651 \\
Qwen2.5-1.5B     & StrategyQA & 0.432 & 0.647 & 0.824 & 0.901 & 0.271 & 0.817 & 0.649 \\
Qwen2.5-1.5B     & Synthetic  & 0.704 & 0.672 & 0.767 & 0.840 & 0.482 & 0.890 & 0.726 \\
\midrule
Phi-2            & GSM8K      & 0.240 & 0.261 & 0.572 & 0.762 & 0.201 & 0.868 & 0.484 \\
Phi-2            & MMLU       & 0.284 & 0.455 & 0.578 & 0.755 & 0.255 & 0.758 & 0.514 \\
Phi-2            & StrategyQA & 0.524 & 0.649 & 0.634 & 0.974 & 0.479 & 0.831 & 0.682 \\
Phi-2            & Synthetic  & 0.640 & 0.591 & 0.618 & 0.936 & 0.578 & 0.831 & 0.699 \\
\bottomrule
\end{tabular}%
}
\end{table}

\textbf{StrategyQA.} 
Phi-2 achieves $\text{LS}=0.974$, the highest LS
across any model--dataset combination, essentially unchanged from the
uncorrected run since LS does not depend on answer extraction, while
LLaMA-3.1-70B drops to $\text{RS}=0.783$ and, more strikingly, to
$\text{CS}=0.287$, the lowest consistency observed for any model on any dataset:
LLaMA-3.1-70B frequently produces a different implicit yes/no judgment
across the $K=3$ resamplings of the same StrategyQA item. ES is notably
low for most models (0.220--0.479) but reaches $0.606$ for
Gemini-2.5-Flash. Because ES is normalized within each model, this value
reflects Gemini-2.5-Flash's observed correctness--token profile and should
not be attributed directly to its larger \texttt{max\_tokens} setting
(Section~\ref{sec:setup}). More generally, commonsense questions tend to
elicit longer traces without proportional correctness gains.

\textbf{Synthetic.} 
CS is more narrowly clustered and elevated here than
on any other dataset (0.591--0.817 across all seven models), reflecting
the dataset's structured, easily-canonicalized answer format. Most API
models maintain RS $>0.85$ (GPT-4o-mini 0.868, DeepSeek-V3 0.935,
Gemini-2.5-Flash 0.880, LLaMA-3.1-70B 0.868), though Claude-Haiku-4.5
dips to RS $=0.749$ on this dataset specifically; local models show
greater vulnerability still (Qwen2.5-1.5B RS $=0.767$, Phi-2
RS $=0.618$).

\subsection{Cross-Cutting Findings}

Three findings emerge consistently across all datasets. First,
CQ and ES are more tightly coupled than most other dimension
pairs, although they can still diverge at the model level: 
Claude-Haiku-4.5
leads on both CQ ($0.783$) and ES ($0.643$), DeepSeek-V3 achieves
comparable CQ ($0.719$, a 6.4-point gap), yet ES falls to $0.540$, a
10.3-point efficiency gap disproportionate to the correctness
difference, and Gemini-2.5-Flash inverts the pattern entirely, which means a lower
CQ ($0.674$) than DeepSeek-V3 but a higher ES ($0.582$). Second,
logical coherence shows no statistically significant association
with correctness under the corrected protocol
: the CQ--LS correlation is $r=-0.083$ (\textit{ns},
$n=28$), consistent with prior findings in the faithfulness
literature~\cite{lanham2023measuring,barez2025chain}, showing
that correct answers can arise from incoherent reasoning traces. LS is,
in fact, the only dimension for which no statistically significant
correlation with any other dimension was observed in the panel
(Section~\ref{sec:discval}). Third,
small models exhibit non-trivial dimensional profiles:
Phi-2 achieves the panel's highest LS ($0.860$) and a competitive
SS ($0.823$) despite the panel's lowest CQ ($0.426$), indicating that
coherence and stability are not simply proxies for correctness, even at
the 2.7B scale.

\subsection{Deployment-Aware Model Selection}
\label{sec:deployment}

These dimension-level differences motivate a practical question: do
alternative deployment priorities materially alter model rankings?
Standard benchmark evaluation produces a single global ranking;
the appropriate model depends on which quality dimensions are most
consequential. Table~\ref{tab:rankings} presents model rankings
across all seven evaluation scenarios.

\begin{table}[H]
\centering
\caption{Model rankings under all seven evaluation scenarios.
Claude-Haiku-4.5 ranks first and DeepSeek-V3 second in every scenario
except Edge Device/IoT. Rankings are descriptive; small score differences
should not be interpreted as statistically reliable because formal
between-model significance tests were not conducted.}
\label{tab:rankings}
\resizebox{\textwidth}{!}{%
\begin{tabular}{L{3.0cm}C{1.2cm}C{1.5cm}C{1.5cm}C{1.2cm}C{1.2cm}C{1.1cm}C{1.1cm}}
\toprule
\textbf{Scenario} & \textbf{\#1} & \textbf{\#2} & \textbf{\#3} & \textbf{\#4} & \textbf{\#5} & \textbf{\#6} & \textbf{\#7} \\
\midrule
Balanced            & Claude & DeepSeek & Gemini   & LLaMA & GPT & Qwen  & Phi-2 \\
Safety Priority     & Claude & DeepSeek & Gemini   & LLaMA & GPT & Qwen  & Phi-2 \\
Accuracy Priority   & Claude & DeepSeek & Gemini   & LLaMA & GPT & Qwen  & Phi-2 \\
Efficiency Priority & Claude & DeepSeek & Gemini   & LLaMA & GPT & Qwen  & Phi-2 \\
Medical Triage      & Claude & DeepSeek & Gemini   & LLaMA & GPT & Qwen  & Phi-2 \\
Legal/Compliance    & Claude & DeepSeek & Gemini   & LLaMA & GPT & Qwen  & Phi-2 \\
Edge Device/IoT     & Claude & Gemini   & DeepSeek & LLaMA & GPT & Phi-2 & Qwen  \\
\bottomrule
\end{tabular}%
}
\end{table}

\rev{Rankings are highly stable across scenarios under the
corrected protocol.
Unlike the substantial rank inversions observed
under the earlier protocol, the corrected results produce the same
ordering for the top five models (Claude-Haiku-4.5, DeepSeek-V3,
Gemini-2.5-Flash, LLaMA-3.1-70B, and GPT-4o-mini) across all seven weighting
schemes. This
stability is a direct, expected consequence of the discriminant-validity
finding in Section~\ref{sec:discval}: because CQ, CS, RS, and ES are now
strongly intercorrelated ($r=0.57$--$0.93$) once computed under a shared
canonicalization protocol, no weighting scheme built from these four
dimensions can substantially reorder models that differ mainly along the
correlated axis. Ranking sensitivity is, therefore, most likely to arise
in scenarios that assign greater weight to the comparatively independent
trace-level dimensions, LS and SS, or that strongly amplify model-level
differences in ES.}

Key inversion: Gemini-2.5-Flash vs.\ DeepSeek-V3, and
Phi-2 vs.\ Qwen2.5-1.5B, under Edge Device/IoT. 
Edge Device/IoT is the only scenario in which the ranking departs from
the otherwise-universal ordering, and it produces two simultaneous rank
swaps. First, Gemini-2.5-Flash (Edge score $0.662$) overtakes
DeepSeek-V3 ($0.657$) to move into second place, despite DeepSeek-V3 leading Gemini
on CQ ($0.719$ vs.\ $0.674$) and RS ($0.927$ vs.\ $0.914$): Gemini's
efficiency advantage (ES $=0.582$ vs.\ $0.540$) is amplified by the
scenario's 50\% ES weight, which is enough to offset its correctness deficit.
Second, among the two smallest models, Phi-2 (Edge score $0.460$)
overtakes Qwen2.5-1.5B ($0.459$), the only scenario in which Phi-2
outranks Qwen2.5-1.5B, and this is driven by the same mechanism: Phi-2 trails
Qwen2.5-1.5B on four of six dimensions (CQ, CS, RS, and SS) but leads on LS
and, decisively for this scenario, ES ($0.395$ vs.\ $0.328$). Both
inversions illustrate the same underlying point: a deployment context
that heavily weights efficiency can favor a model that is otherwise
weaker on accuracy-adjacent dimensions, and this preference is invisible
to any correctness-only benchmark. A practitioner relying on
accuracy-only benchmarking would rank DeepSeek-V3 and Qwen2.5-1.5B ahead
of their Edge-appropriate alternatives in precisely the deployment
context of resource-constrained inference, where that ranking is
least appropriate.

\subsection{Discriminant Validity}
\label{sec:discval}

\begin{figure}[H]
  \includegraphics[width=\textwidth]{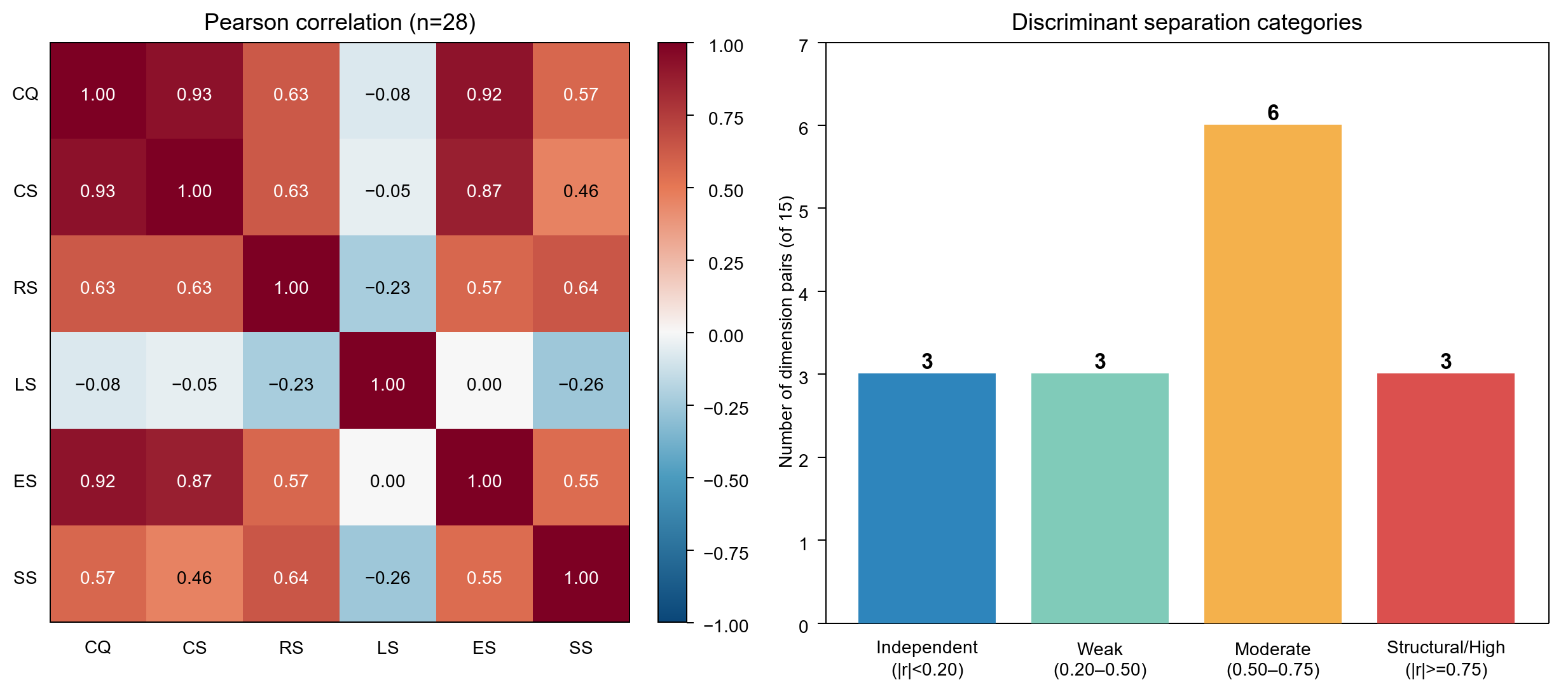}
  \caption{Discriminant 
 validity visualization. (\textbf{Left}) Pearson
  correlation heatmap ($n=28$). (\textbf{Right})~Categorical classification
  of dimension pairs. Structural correlations are highlighted separately
  to distinguish them from problematic metric overlap.}
  \label{fig:discval}
\end{figure}

The observed stability of most rankings raises a broader measurement
question: to what extent do the six dimensions capture empirically
distinct constructs? Discriminant validity (evidence that the proposed dimensions measure
non-overlapping constructs) is a necessary condition for any
multi-dimensional framework. We report Pearson correlations with 95\%
bootstrap confidence intervals over $n = 28$ observations (7 models
$\times$ 4 datasets). Because these observations reuse the same model
families and benchmark domains, they are not fully independent in the
psychometric sense, and the results should, therefore, be interpreted as
indicative rather than conclusive construct-validity evidence.
Figure~\ref{fig:discval} and Table~\ref{tab:discval} summarize the
results.

\rev{Of the 15 dimension pairs, 6 have negligible or weak correlations
($|r|<0.50$), whereas the remaining 9 are elevated because CQ, CS, RS, and ES share the same canonicalized answer representation or correctness signal, whether
directly or through correctness-conditioning
(Section~\ref{sec:setup}): a model that produces cleanly extractable,
correct answers tends to score well on all four simultaneously.
CQ--ES ($r=0.917$) and CQ--RS ($r=0.627$) remain structural by
construction, as in the original design (RS is defined only over
correctly-answered items and ES's harmonic mean embeds $CQ_i$ directly). Unlike CQ--ES and CQ--RS, the
newly prominent association between CQ--CS 
($r=0.928$) is not
definitionally dependent in the same sense (CS is computed
independently, as cross-run agreement among $K=3$ samples) but is
empirically driven by the shared canonicalization step: once both metrics compare against the
same extracted-answer space, a model either reliably lands on the
canonical answer (high CQ and CS together) or does not. Partial
correlation confirms CQ-mediation across the board: controlling for CQ
collapses RS--ES from $r=0.566$ to $r_{\text{RS--ES}\mid\text{CQ}}=-0.029$,
CS--ES from $r=0.874$ to $r_{\text{CS--ES}\mid\text{CQ}}=0.153$, and
CS--RS from $r=0.634$ to $r_{\text{CS--RS}\mid\text{CQ}}=0.179$;
in each case the raw association is almost entirely attributable to a
shared dependence on CQ rather than to genuine pairwise redundancy.}

\begin{table}[H]
\centering
\caption{Pearson correlations between all 15 dimension pairs ($n=28$),
recomputed under the corrected canonical-answer-extraction protocol.
$^{***}p<0.001$, $^{**}p<0.01$, $^{*}p<0.05$, and \textit{ns} = not significant.
\checkmark\ = negligible and not statistically significant; $\sim$\ = weak;
!\ = moderate or higher; and $\triangle$\ = structural (definitional or
shared-pipeline).}
\label{tab:discval}
\resizebox{\textwidth}{!}{%
\begin{tabular}{C{1.8cm}C{1.2cm}C{1.4cm}C{1.0cm}C{3.4cm}L{6cm}}
\toprule
\textbf{Pair} & $\mathbf{r}$ & \textbf{\em{p}} & \textbf{Sig.} & \textbf{95\% CI} & \textbf{Interpretation} \\
\midrule
CQ--CS & $+0.928$ & $<0.001$ & $***$       & $[+0.864,+0.972]$ & $\triangle$\ Shared extraction pipeline \\
CQ--RS & $+0.627$ & $<0.001$ & $***$       & $[+0.321,+0.815]$ & $\triangle$\ Structural (definitional) \\
CQ--LS & $-0.083$ & $0.676$  & \textit{ns} & $[-0.435,+0.291]$ & \checkmark\ Negligible; no significant association \\
CQ--ES & $+0.917$ & $<0.001$ & $***$       & $[+0.832,+0.964]$ & $\triangle$\ Structural (definitional) \\
CQ--SS & $+0.573$ & $0.001$  & $**$        & $[+0.320,+0.754]$ & !\ Moderate \\
CS--RS & $+0.634$ & $<0.001$ & $***$       & $[+0.400,+0.805]$ & !\ Moderate (CQ-mediated) \\
CS--LS & $-0.054$ & $0.785$  & \textit{ns} & $[-0.413,+0.324]$ & \checkmark\ Negligible; no significant association \\
CS--ES & $+0.874$ & $<0.001$ & $***$       & $[+0.744,+0.956]$ & !\ Moderate--high (CQ-mediated) \\
CS--SS & $+0.458$ & $0.014$  & $*$         & $[+0.242,+0.645]$ & $\sim$\ Weak \\
RS--LS & $-0.229$ & $0.241$  & \textit{ns} & $[-0.586,+0.157]$ & $\sim$\ Weak \\
RS--ES & $+0.566$ & $0.002$  & $**$        & $[+0.269,+0.778]$ & !\ Moderate (CQ-mediated) \\
RS--SS & $+0.637$ & $<0.001$ & $***$       & $[+0.321,+0.821]$ & !\ Moderate \\
LS--ES & $+0.001$ & $0.995$  & \textit{ns} & $[-0.336,+0.360]$ & \checkmark\ Negligible; no significant association \\
LS--SS & $-0.262$ & $0.178$  & \textit{ns} & $[-0.638,+0.119]$ & $\sim$\ Weak \\
ES--SS & $+0.548$ & $0.003$  & $**$        & $[+0.278,+0.750]$ & !\ Moderate \\
\bottomrule
\end{tabular}%
}
\end{table}

\rev{Read this way, the six-dimensional framework's clearest discriminant
contribution is concentrated in one dimension: LS 
shows no
statistically significant association with any other dimension in the
panel (all
$|r|\leq0.229$, none significant), including, notably, with CQ
($r=-0.083$), which is consistent with prior findings in the faithfulness literature
\cite{lanham2023measuring,barez2025chain} that correct answers can arise
from incoherent reasoning traces even when CQ, CS, RS, and ES move
together. SS 
retains partial independence, correlating weakly
with CS ($r=0.458$) and moderately with the CQ-cluster
($r=0.55$--$0.64$) but never approaching the near-unity association seen
within that cluster itself, which is consistent with our earlier finding
(Section~\ref{sec:results}) that the semantic content of reasoning
traces is more stable across runs than the exact extracted~answer.}

\begin{figure}[H]
  \includegraphics[width=0.85\textwidth]{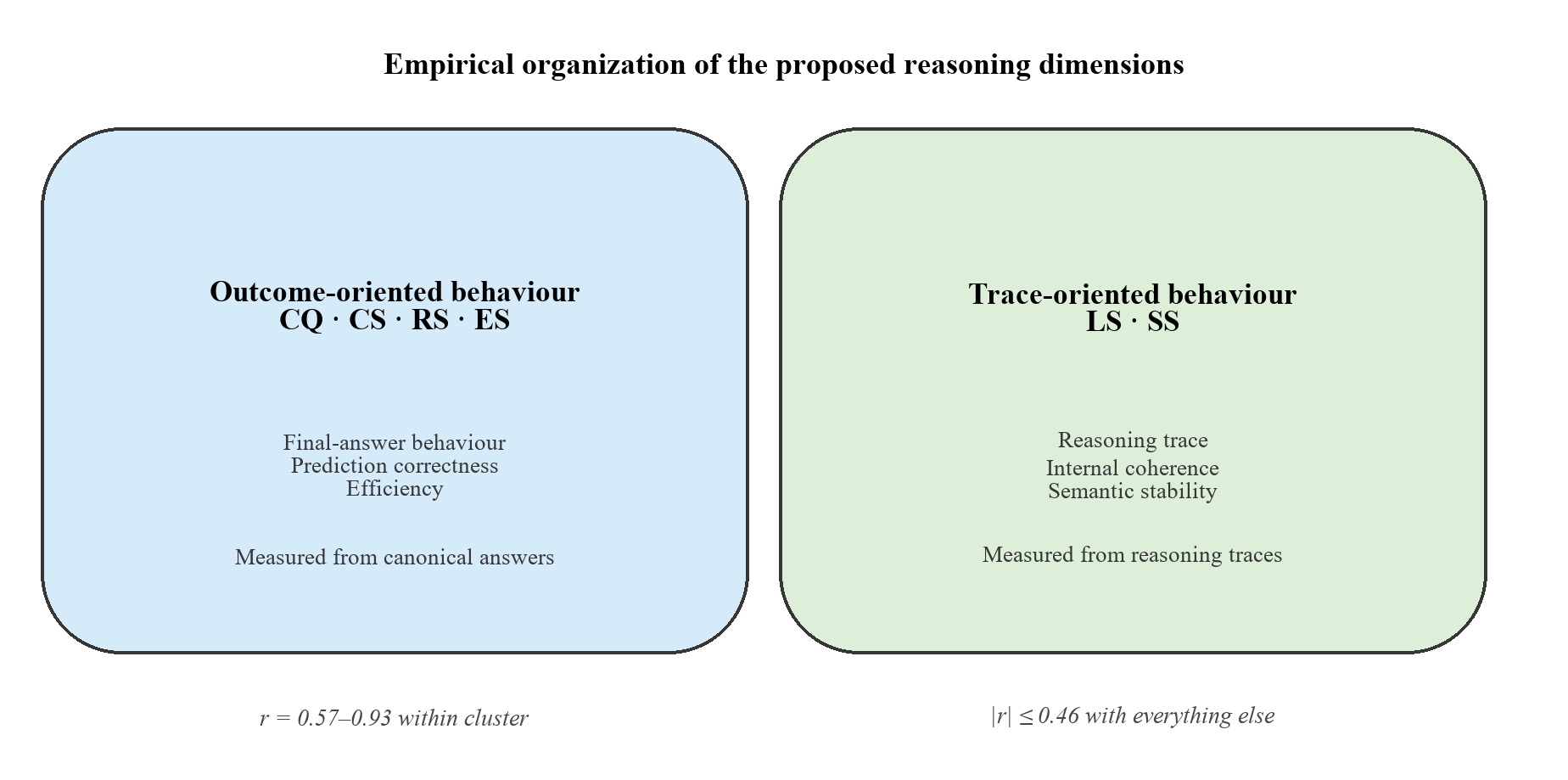}
  \caption{Empirical 
organization of the proposed reasoning dimensions.
Outcome-oriented metrics (CQ, CS, RS, and ES) are measured from
canonicalized final answers and are, consequently, empirically
intercorrelated ($r=0.57$--$0.93$). Trace-oriented metrics (LS and SS)
are measured from reasoning traces: LS shows no statistically significant
association with the outcome metrics, whereas SS retains moderate
associations with several of them; LS and SS are not significantly
correlated with each other. The blue panel denotes the outcome-oriented
layer, and the green panel denotes the trace-oriented layer. The two panels
summarize the empirical organization of the measured reasoning
dimensions, not a causal or temporal relationship.}
  \label{fig:twolayer}
\end{figure}

\rev{Figure~\ref{fig:twolayer} summarizes the principal empirical
finding of this study. Although the framework defines six behavioral
dimensions on independent theoretical grounds (Section~\ref{sec:overview}), their empirical organization across the seven evaluated
models reveals two behavioral layers: outcome-level metrics are
linked through canonicalized final answers or correctness signals and,
therefore, exhibit relatively strong pairwise correlations, whereas
trace-level metrics are derived directly from reasoning traces. Within the
trace-level layer, LS is the more clearly distinct dimension, while SS
retains moderate associations with several outcome-level metrics. This
separation explains both the discriminant-validity
structure reported above and the deployment behavior reported in
Section~\ref{sec:deployment}.}

\rev{In summary, the framework's strongest orthogonal information is carried
by LS, while SS provides a partially distinct trace-level signal; this
information is not distributed uniformly across all six dimensions.} The 95\% CIs were
confirmed via bootstrap validation ($B=10{,}000$~resamples).

\section{Discussion}
\label{sec:discussion}

\subsection{Practical Implications}
\label{sec:implications}

The discriminant validity results (Section~\ref{sec:discval}) point to a
single organizing pattern that runs through the findings below: the six
dimensions do not behave as six uniformly independent signals but
rather as two behavioral layers. An \emph{outcome-level} cluster (Correctness,
Consistency, Robustness, and Efficiency) is strongly intercorrelated
($r=0.57$--$0.93$) because all four share a canonicalized answer
representation or correctness signal. At the \emph{trace level}, Local
Logical Coherence shows no statistically significant association with the
outcome-level metrics, whereas Stability retains moderate associations with
several of them; LS and SS are not significantly correlated with each
other. The practical implications below follow from this structure.

\textbf{Implication 1: Local logical coherence contributes information
beyond final-answer correctness.} 
In the evaluated sample, LS showed no
statistically significant association with any other dimension, including
CQ ($r=-0.083$, \textit{ns}). It, therefore, captures a property that is not
directly represented by a correctness score. Phi-2 on StrategyQA
illustrates this diagnostic distinction: $\text{CQ}=0.524$ and
$\text{LS}=0.974$ show that relatively high local coherence can coexist
with moderate final-answer correctness. This pattern does not, by itself,
identify the cause of the errors, because LS detects local contradiction
rather than knowledge deficiencies, causal faithfulness, or global
argument validity. For interpretability-sensitive systems (such as
legal assistants and clinical decision support), a reasoning trace
should, therefore, be evaluated separately from the correctness of its final
answer.

\textbf{Implication 2: Within the outcome-level cluster, model rankings
are largely stable across deployment weightings, with one informative
exception.} 
Because CQ, CS, RS, and ES move together, weighting schemes
built from these four dimensions rarely disagree (Table~\ref{tab:rankings}). The exception is the Edge Device/IoT
scenario, which assigns an unusually large weight to efficiency (50\%)
and thereby amplifies model-level differences within the
outcome-oriented cluster: Gemini-2.5-Flash overtakes DeepSeek-V3 despite trailing it
on correctness and robustness, and Phi-2 overtakes Qwen2.5-1.5B despite
trailing it on four of six dimensions (Section~\ref{sec:deployment}). A
procurement process optimizing for accuracy alone could, therefore, favor a
different model in a resource-constrained setting. These reversals are
descriptive and based on small score differences, but they illustrate how
deployment weights can alter a decision when efficiency is prioritized.

\textbf{Implication 3:
The efficiency--correctness gap remains
practically meaningful, even within the correlated cluster.} DeepSeek-V3
($\text{CQ}=0.719$) reaches $\text{ES}=0.540$ versus Claude-Haiku-4.5's
$\text{ES}=0.643$ (a 10.3-point efficiency gap against a 6.4-point
correctness difference), and Gemini-2.5-Flash inverts the pattern
outright, trailing DeepSeek-V3 on correctness yet leading it on
efficiency. A strong aggregate correlation does not imply that the
dimensions are redundant \emph{for every individual pair of models}; it
means that, in aggregate, a practitioner who observes CQ can often
predict CS, RS, and ES reasonably well, but not always. The exceptions may
be particularly relevant to deployment decisions.

Taken together, these implications reframe the framework's contribution;
it is not merely a ranking tool but a \emph{pre-deployment diagnostic instrument}
that is most informative along two axes:  whether a model's reasoning
is internally coherent (LS), and whether its accuracy-adjacent behavior
(CQ, CS, RS, and ES) diverges from the norm for its evaluated cluster
peers, as it does for Gemini-2.5-Flash and Phi-2 under efficiency
weighting.

\subsection{Limitations}
These implications should be interpreted in light of several
limitations.
First, the discriminant validity analysis is based on $n = 28$
model--dataset observations that are not fully independent,
and significance tests should be interpreted accordingly.
Second, model rankings across deployment scenarios are
reported as point estimates; between-model score differences
are not accompanied by significance tests, and small gaps
(e.g., $\Delta Q < 0.01$) should not be over-interpreted.
Third, the LS metric captures local step-to-step contradiction
via NLI and does not assess causal faithfulness or global
argument validity. Fourth, all evaluations were conducted
at temperature $= 0.7$; consequently, CS partly reflects the selected
sampling regime rather than intrinsic model inconsistency alone.
Replication at lower temperatures is, therefore, warranted. Future versions of the LS metric may incorporate
reasoning-depth normalization to avoid rewarding
trivially short responses with perfect scores.

\rev{Fifth, specifically in the reproduced analysis, the corrected
canonical-answer-extraction protocol (Section~\ref{sec:setup})
substantially increased the empirical correlation between CQ, CS, RS,
and ES (Section~\ref{sec:discval}). This is, in our reading, a more
accurate picture of these metrics' relationships than the one obtained
under the original full-text CS computation, but it also means the
framework's discriminant-validity evidence is now concentrated in fewer
dimensions (principally LS) than originally claimed; readers should
weight the six dimensions accordingly rather than treating each as
equally informative. Sixth, one model in our panel
(Gemini-2.5-Flash) required a non-default \texttt{max\_tokens} setting
(1024 vs.\ 256 for all other API models) to avoid the systematic truncation
caused by the gateway's internal reasoning-token consumption; because ES
is computed via min--max normalization over the token counts observed
within a given model's own evaluation run (Table~\ref{tab:notation}),
this does not bias Gemini-2.5-Flash's own ES score, but readers should
not interpret the raw token counts underlying ES as directly comparable
generation-cost figures across models with different provider-side
token-budget behavior. Seventh, we identified and corrected a
transcription error in two of the seven built-in aggregation weight
vectors (Safety Priority and Accuracy Priority; Table~\ref{tab:weights})
between the originally released configuration file and the manuscript's
reported values; all results in this paper use the corrected weights,
which are also the values shipped in the released code. Eighth, CS (and,
by the same argument, CQ and RS) is not directly comparable in raw
magnitude across datasets with different answer-space cardinalities. A
model that reproduces the same answer by chance is far more likely to do
so on a binary yes/no item (StrategyQA) than on an open-ended numeric
item (GSM8K): under a naive uniform-random-guessing baseline, two
independent samples agree with probability $0.50$ for a binary answer
space but approach $0$ for an unconstrained numeric one. The
per-dataset CS values reported in Table~\ref{tab:perdataset} (e.g.,
GPT-4o-mini: $\text{CS}=0.531$ on StrategyQA vs.\ $\text{CS}=0.567$ on
GSM8K) should, therefore, be read as within-dataset comparisons across
models, not as evidence that a model is equally consistent in an
absolute sense across answer types; a chance-corrected variant of CS
(e.g., Cohen's $\kappa$-style adjustment for expected agreement under
the answer space's cardinality) is a natural refinement for future
versions of the framework and would be required before CS scores are
compared in magnitude across datasets of differing answer-space size.}

\section{Conclusions}
\label{sec:conclusion}

This paper proposed a multi-dimensional behavioral framework that
operationalizes six complementary dimensions of LLM reasoning quality in a
unified and reproducible evaluation pipeline. The empirical results indicate
that these dimensions form two behavioral layers rather than six equally
independent axes. The outcome-oriented dimensions (Correctness,
Consistency, Robustness, and Efficiency) are strongly intercorrelated
because they share a canonicalized answer representation or correctness
signal. Within the trace-level layer, Local Logical Coherence showed no
statistically significant association with any other dimension in the
evaluated sample, whereas Stability retained moderate associations with
several outcome-level metrics. These findings suggest that LS provides
information not directly captured by final-answer evaluation.
Efficiency-weighted scenarios produced limited ranking changes when
model-level efficiency differences received substantial weight.

\textls[-15]{The framework's deployment-aware aggregation mechanism translates
dimensional profiles into context-specific decision support, enabling
practitioners to move beyond generic leaderboard rankings. The discriminant
validity analysis provides preliminary evidence that local logical coherence
contributes a non-redundant signal in the evaluated sample. The structure of
the outcome-level cluster and the extent to which it generalizes across model
families remain important questions for validation on domain-specific~benchmarks.}

RQEval is model-agnostic, requires no access to model weights, and is publicly available to support reproducibility and further development by the research community. Overall, these findings suggest that future reasoning benchmarks should distinguish between outcome-oriented and trace-oriented behavioral properties rather than relying on correctness alone.

\vspace{6pt}





\authorcontributions{Conceptualization, A.Ş. and H.L.; methodology, A.Ş.; software, A.Ş.; validation, A.Ş., G.A. and H.L.; formal analysis, A.Ş.; investigation, A.Ş.; data curation, A.Ş.;
writing---original draft preparation, A.Ş.; writing---review and editing, G.A. and H.L.; visualization, A.Ş.; and supervision, H.L. All authors have read and agreed to the published version of the manuscript.}

\funding{
This research received no external funding.}

\dataavailability{The following datasets used in this study are publicly available: GSM8K~\cite{cobbe2021training}, MMLU~\cite{hendryckstest2021}, and StrategyQA~\cite{geva-etal-2021-aristotle}. The Synthetic dataset is distributed with the framework code at  
\url{https://github.com/senolali/RQEval} (accessed on 23 July 2026).
The framework is also available as a Python 3.13 
 package at
\url{https://pypi.org/project/RQEval/} and can be
installed using \texttt{pip install rqeval}.}

\acknowledgments{The authors thank Chengshuai Zhao (Arizona State University)
for his valuable feedback on an earlier draft of this manuscript.}
%
%

\conflictsofinterest{The authors declare no conflicts of interest.}





\newpage
\abbreviations{Abbreviations}{
			The following abbreviations are used in this manuscript:
			\\

			\noindent
			\begin{tabular}{@{}ll}
				LLM  & Large Language Model \\
                CQ   & Correctness Quality \\
                CS   & Consistency Score \\
                RS   & Robustness Score \\
                LS   & Local Logical Coherence Score \\
                ES   & Efficiency Score \\
                SS   & Stability Score \\
                CoT  & Chain-of-Thought \\
                NLI  & Natural Language Inference \\
                MNLI & Multi-Genre Natural Language Inference \\
                GSM8K & Grade School Math 8K \\
                MMLU & Massive Multitask Language Understanding \\
                API  & Application Programming Interface \\
                VRAM & Video Random Access Memory \\
NLP  & Natural Language Processing \\
			\end{tabular}
		}


\reftitle{References}




\begin{adjustwidth}{-\extralength}{0cm}


\PublishersNote{}
\end{adjustwidth}

%



\end{document}